\documentclass[letterpaper]{article} % DO NOT CHANGE THIS
\pdfoutput=1
\usepackage{aaai23}  % DO NOT CHANGE THIS
\usepackage{times}  % DO NOT CHANGE THIS
\usepackage{helvet}  % DO NOT CHANGE THIS
\usepackage{courier}  % DO NOT CHANGE THIS
\usepackage[hyphens]{url}  % DO NOT CHANGE THIS
\usepackage{graphicx} % DO NOT CHANGE THIS
\usepackage{natbib}  % DO NOT CHANGE THIS AND DO NOT ADD ANY OPTIONS TO IT
\usepackage{caption} % DO NOT CHANGE THIS AND DO NOT ADD ANY OPTIONS TO IT
\usepackage{subfigure}
\usepackage{algorithm}
\usepackage{algorithmic}
\usepackage{multirow}
\usepackage{booktabs}
\usepackage{amsmath}
\usepackage{amsfonts}

\usepackage{newfloat}
\usepackage{listings}
\DeclareCaptionStyle{ruled}{labelfont=normalfont,labelsep=colon,strut=off} % DO NOT CHANGE THIS
\floatstyle{ruled}
\newfloat{listing}{tb}{lst}{}
\floatname{listing}{Listing}
\title{STOA-VLP: Spatial-Temporal Modeling of Object and Action for Video-Language Pre-training
}
\author {
    Weihong Zhong\textsuperscript{\rm 1},
    Mao Zheng\textsuperscript{\rm 2},
    Duyu Tang\textsuperscript{\rm 3},
    Xuan Luo\textsuperscript{\rm 2},
    Heng Gong\textsuperscript{\rm 1}, \\
    Xiaocheng Feng\textsuperscript{\rm 1,4}\thanks{Corresponding Author},
    Bing Qin\textsuperscript{\rm 1,4}
}
\affiliations {
    \textsuperscript{\rm 1} Harbin Institute of Technology\\
    \textsuperscript{\rm 2} Tencent MLPD\\
    \textsuperscript{\rm 3} Independent Researcher\\
    \textsuperscript{\rm 4} Peng Cheng Laboratory\\
    \{whzhong, hgong, xcfeng, qinb\}@ir.hit.edu.cn, \{moonzheng, ttssxuanluo\}@tencent.com, \\ tangduyu.nlp@hotmail.com
}
\begin{document}

\maketitle

\begin{abstract}
Although large-scale video-language pre-training models, which usually build a global alignment between the video and the text, have achieved remarkable progress on various downstream tasks, the idea of adopting fine-grained information during the pre-training stage is not well explored.
In this work, we propose STOA-VLP, a pre-training framework that jointly models object and action information across spatial and temporal dimensions. 
More specifically, the model regards object trajectories across frames and multiple action features from the video as fine-grained features.
Besides, We design two auxiliary tasks to better incorporate both kinds of information into the pre-training process of the video-language model.
The first is the dynamic object-text alignment task, which builds a better connection between object trajectories and the relevant noun tokens. 
The second is the spatial-temporal action set prediction, which guides the model to generate consistent action features by predicting actions found in the text. 
Extensive experiments on three downstream tasks (video captioning, text-video retrieval, and video question answering) demonstrate the effectiveness of our proposed STOA-VLP (e.g. 3.7 Rouge-L improvements on MSR-VTT video captioning benchmark, 2.9\% accuracy improvements on MSVD video question answering benchmark, compared to previous approaches). 

\end{abstract}
\section{Introduction}
\label{intro}
\begin{figure}[t]
\centering
\includegraphics[width=1.0\columnwidth]{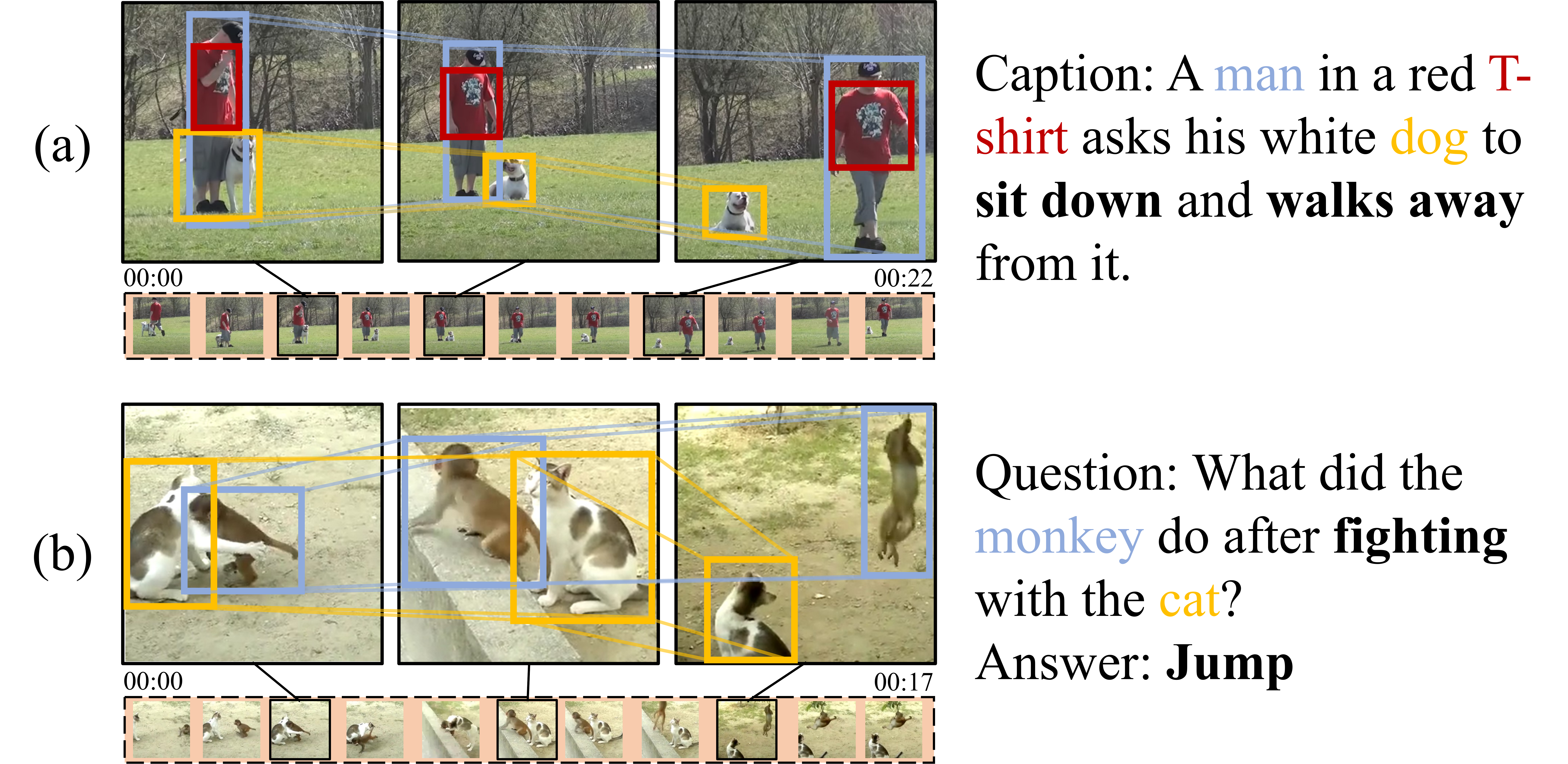} % Reduce the figure size so that it is slightly narrower than the column. Don't use precise values for figure width.This setup will avoid overfull boxes.
\caption{(a): A video captioning example. Recognitions of salient objects and movements of objects are important clues for generating the caption. (b): A video question answering example. To answer this question, a fine-grained connection between salient objects should be built, as well as object movements and their different interactions should be carefully analyzed.}

\label{actionillu}
\end{figure}

With the introduction of large-scale datasets, video-language pre-training methods learn to build adaptive cross-modal representations for downstream video-text understanding and generation tasks. 
Previous dominant approaches focus on learning an overall alignment from raw video and text data \cite{lei2021less, bain2021frozen}, lacking modeling of fine-grained information. Meanwhile, fine-grained information can be necessary for understanding the video scene in many situations.
For example, as Figure \ref{actionillu}(a) depicts, generating such captions requires detailed dynamic modeling of relevant objects and movements across time. To answer questions like Figure \ref{actionillu}(b), a fine-grained object-text alignment is required, as well as object movements and their different interactions should be carefully analyzed. 
A coarse-grained method can easily fail at handling such information. 

\begin{figure*}[t]
\centering
\includegraphics[width=2.0\columnwidth]{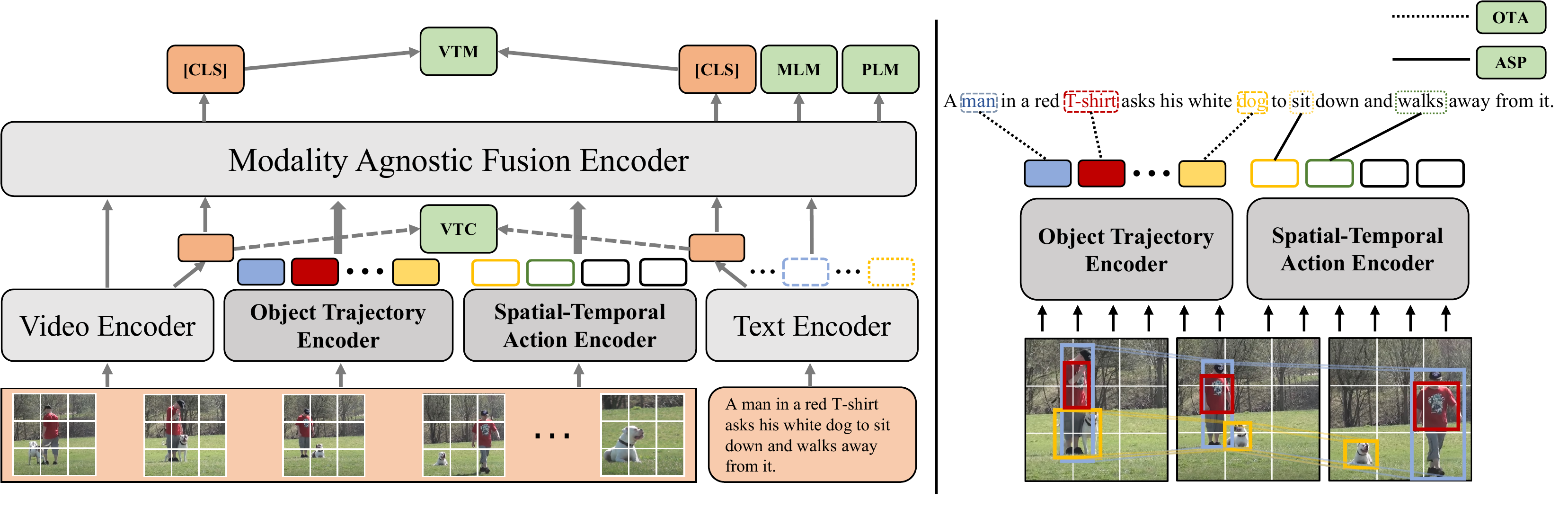} % Reduce the figure size so that it is slightly narrower than the column. Don't use precise values for figure width.This setup will avoid overfull boxes.
\caption{Left: Overview of our STOA-VLP's architecture. STOA-VLP contains a video encoder, a text encoder, as well as two newly proposed encoders: the object trajectory encoder and the spatial-temporal action encoder, which are used to explicitly model dynamic object trajectories and action information, respectively. On top of them, a modality-agnostic fusion encoder is built to enable further cross-modal interaction. Right: We design two auxiliary tasks to build a fine-grained alignment with the text. In the object-text alignment (OTA) task, we align object trajectories with noun tokens in the paired text. In the action set prediction (ASP) task, we predict actions found in the text with generated action features to guide the training of the action encoder. }

\label{model}
\end{figure*}

Efforts have been made by previous pre-training approaches to maintain fine-grained information from different aspects. Although achieved promising improvements, these methods still have some limitations: (1) Objects are usually moving and changing across time. Existing methods \cite{li2022align, wang2022object} usually utilize fixed bounding boxes to model objects, leading to an insufficient understanding of object movements in the video. (2) There can be multiple overlapping actions at the same time range, triggered by different object interactions in different spatial regions. Existing methods \cite{zhu2020actbert} focus on only one action representation for each video clip, which leads to an information loss. 
As Figure \ref{actionillu} shows, the main content of the video can be described by the movement of objects and the interaction between them. 
Therefore, an ideal approach is to incorporate these two kinds of information into the pre-training stage for fine-grained spatial-temporal modeling.

In this paper, we tackle the above problems by proposing a new pre-training framework---STOA-VLP: Spatial-Temporal Modeling of Object and Action for Video-Language Pre-training that captures and jointly models object trajectories and multiple actions to better handle movements and interactions of objects in the video.
More specifically, we automatically extract bounding boxes from multiple frames. Temporal-aware object trajectories are generated using raw video patches and bounding boxes across frames with identical labels.
Next, raw video patches and bounding boxes from different frames are utilized to generate multiple spatial-temporal action features with different action query tokens. In addition, to better incorporate object trajectories and action features, two novel auxiliary tasks are designed to build a fine-grained alignment with the text (see the right of Figure \ref{model}). 
The first one is dynamic object-text alignment (OTA), which calculates an optimal bipartite matching between generated object trajectories and noun tokens in the paired text and maximizes the matching score to dynamically connect the object feature with the paired text. 
The second one is spatial-temporal action set prediction (ASP), which firstly constructs an action set out of the text, then constructs a prediction matrix with action features to find an optimal bipartite matching between them. 
The matching score is maximized to enhance their alignment.

Finally, we evaluate our STOA-VLP on three commonly used tasks including video captioning, text-video retrieval, and video question answering. Our framework achieves substantial improvements over other methods on most of the metrics with only 2.5M pre-training data. To sum up, our contributions are:

\begin{itemize}
\item We propose STOA-VLP that explicitly models temporal-aware object trajectories and multiple spatial-temporal action features to better handle the dynamic movements and interactions of objects in the video.\footnote{Our code will be available at https://github.com/whongzhong/
STOA-VLP.}

\item We design two novel auxiliary pre-training tasks---OTA and ASP to better model object trajectories and action features during the pre-training stage. 

\item Our STOA-VLP surpasses previous approaches on most of the metrics, and achieves a large performance gain over our base model, for example, achieving 1.9 CIDEr improvements on the MSR-VTT video captioning benchmark and 1.2\% absolute accuracy improvements on the MSVD video question answering benchmark.

\end{itemize}
 
\section{Related Work}
\subsection{Video-Language Pre-training}
Earlier works usually adopt a dense sampling way to extract consecutive frames. Due to the limit of computational cost when extracting offline features from videos \cite{sun2019videobert, sun2019learning,luo2020univl} using ResNet \cite{he2016deep}, S3D \cite{xie2018rethinking} or Faster r-cnn \cite{ren2015faster}, etc., information from consecutive frames tends to be similar, which inspires the idea of sparse sampling. With this thought, \citet{lei2021less} enable the end-to-end pre-training with image-text data and fine-tune with only a few sampled video frames, demonstrating the potential of knowledge transfer from image to video. Furthermore, \citet{bain2021frozen} and \citet{luo2021clip4clip} adopt a dual encoder structure to align video and text with simple instance-level contrastive objectives.
However, different from image data, scenes are changing across video frames, which makes it important to build a temporal alignment. 
As far as temporal information is concerned, \citet{li2020hero} try to model frame-level temporal representations. \citet{zellers2021merlot} and \citet{tang2021decembert} further build an alignment between video clips and temporal marked sentences.

Most of these works focus on learning an overall connection between video and text modality, without explicitly handling fine-grained information. Though achieved advanced results, it can be insufficient when detailed semantic information is needed (e.g. in the video captioning and video question answering tasks).

\subsection{Object and Action Modeling in Video-Language Pre-training}
Since objects and their interactions can describe the major content of a video, fine-grained information considered by most works can fall into two categories, i.e. object and action information. 
(1) To model object information, \citet{wang2022object} and \citet{li2022align} utilize fixed object regions to build a fine-grained video-text alignment, which can be insufficient to handle moving objects and changing scenes. previous video captioning works have tried to construct a graph to model object interactions based on their spatial locations in consecutive frames \cite{zhao2018video, pan2020spatio}, which can be hard when sampled frames are sparse or changing fast. \citet{herzig2022object} propose to implicitly refine the video encoder with object bounding box features without further explicitly modeling inter-frame relationships. 
(2) To model actions, \citet{zhu2020actbert} explicitly train an action recognizer to generate a single action feature from a whole video clip, ignoring the multiple actions triggered by different object interactions. 
\citet{ge2022bridging} take a step further to consider object and action information concurrently. Meanwhile, they construct verb and noun multiple-choice questions, implicitly answering them with the intermediate video tokens to refine the instance-level alignment for a dual-encoder structure. Without further modeling the cross-modal interaction limits its application.

Since objects are moving and interacting across video frames, as well as multiple actions can be triggered by different interactions, those mentioned above are insufficient to fully address semantic information. Our proposed method jointly considers modeling object and action information at both spatial and temporal dimensions. With our designed OTA and ASP tasks, a coarse-to-fine video-text alignment is built to learn a better multi-modality representation.
\section{Method}

In this section, we first introduce the overall model architecture of our STOA-VLP and further discuss the proposed object trajectory encoder and spatial-temporal action encoder. Then, we present the pre-training objectives used to model overall and fine-grained cross-modal alignment, especially our proposed auxiliary tasks OTA and ASP. 
\subsection{STOA-VLP Model Architecture}
\label{overall}
\subsubsection{Overall Architectures}
The overall model structure of our STOA-VLP is shown on the left of Figure \ref{model}, which contains four modality-specific encoders: the video encoder, the text encoder, the object trajectory encoder, and the spatial-temporal action encoder. Firstly, raw video and text data are separately encoded by the video and text encoder. We sparsely extract object bounding box features from the video. The intermediate encoded video features and extracted bounding box features are utilized to generate object trajectories and action features with the corresponding encoders. A modality-agnostic fusion encoder is built on top of these four encoders, in which the overall video and text information, object trajectories, and spatial-temporal-based action information interact with each other. All five modules adopt the  Transformer \cite{vaswani2017attention} structure. We will further introduce each module in order:
\subsubsection{Video Encoder}
Given a video with uniformly sparsely sampled $T$ frames, our video encoder extracts $H \times W$ non-overlapping image patches from each frame and linearly projects them into 1-D tokens. We denote the intermediate patch output as $z \in \mathbb{R}^{T \times HW \times h}$, where $h$ is the dimension of the visual representation. 
Following \cite{dosovitskiy2020image}, we utilize $\left[\mathtt{CLS}\right]$ tokens prepended to each frame as the frame level representation and linearly project them into cross-modal representation space. The final output of video encoder is $ v =\{v_1, v_2, \cdots, v_{T}\}$. where $v_i\in \mathbb{R}^{1 \times d}$ and $d$ is the dimension of the cross-modal representation.
%%%%%
\subsubsection{Text Encoder}

For the given paired text, we tokenize it and add two special tokens $\left[\mathtt{CLS}\right]$ and $\left[\mathtt{SEP}\right]$ to the beginning and end respectively. Assuming that the processed text is of length $L$, we use a text encoder to encode it and acquire the output $t =\{t_1, t_2, \cdots, t_L\}$, where $t_i\in \mathbb{R}^{1 \times d}$.
%%%%%

\subsubsection{Object Trajectory Modeling}

As mentioned in the Introduction, lacking the modeling of objects' movement between frames might result in sub-optimal performance. Therefore, we propose to explicitly model object trajectories between frames to better capture their temporal changing.
Given a sparsely sampled video with $T$ frames, we first extract objects per frame with the off-the-shelf object detector VinVL \cite{zhang2021vinvl}. For each video, We select the top-K distinct objects for frame $\tau$ and keep the bounding box feature (i.e. coordinates) $O^\tau_{\text{box}}\in \mathbb{R}^{K\times 4}$ and object class $O^\tau_{c}\in \mathbb{R}^{K}$. Instead of directly inserting extracted features, We follow a flexible way from \cite{herzig2022object} to get object features via a RoIAlign layer \cite{he2017mask}. With the intermediate video patch output $z \in \mathbb{R}^{T \times HW \times h}$ and bounding box features $O^\tau_{\text{box}}$, we can get object region features of $K$ objects in frame $\tau$ with the following equation: 
\begin{equation}
    O^\tau = \text{MLP}(\text{MaxPool}(\text{RoIAlign}(z, O^\tau_{\text{box}}))),
\end{equation}
getting object features $O = \{O^1, O^2, ..., O^T \} \in \mathbb{R}^{T \times K \times h}$. Note that a spatial-temporal positional embedding is added for each object. To get $N$ object trajectory tokens across all video frames, we sum up the classification confidence for each distinct object label in the $O_c$ within all sampled video frames and select the top-N object class according to the summed confidence. We flatten the extracted object features and get $O_{flatten} \in \mathbb{R}^{(TK) \times h}$. For each selected object class $c$, we generate a class mask $M_c \in \mathbb{R}^{TK}$, the corresponding place of $M_c$ will be set as 0 to be masked if it's not labeled as $c$ in the $O_c$. Therefore, with mask $M_c$, objects labeled as c is visible, while the rest is masked. Despite the fact that this could be a noisy process, We view this mask object sequence as a trajectory for object $c$, since this sequence represents its movements across frames. Similar to our video encoder, we add a $\left[\mathtt{CLS}\right]$ token to the start of each object trajectory sequence. Next, we input the object feature representation $O_{flatten}$ and mask $M_c$ into the object trajectory encoder to model the object trajectory between frames. the output of $\left[\mathtt{CLS}\right]$ is utilized as a single temporal-aware trajectory representation for each selected object class. We denote the final projected output of temporal-aware object representation as $o \in \mathbb{R}^{N \times d}$ for each video. 

\subsubsection{Spatial-Temporal Action Modeling}
The key cues for recognizing actions in videos are: the objects in the scene, their interactions, and their movement \cite{herzig2022object}. Modeling action information naturally requires catching above mentioned clues in the spatial-temporal dimension. Different from \cite{zhu2020actbert} which models a global action for each video clip, we assume there possibly exist multiple actions, and each action can be inferred from different object interactions across frames. Thus, we initialize $M$ action tokens $q = \{q_1, q_2, \cdots,q_M\} \in \mathbb{R}^{M \times h} $ as query to extract action information from videos. Similar to \cite{herzig2022object}, we concatenate the intermediate patch features $z$ with the object token features $O$ to better catch action clues with a local information enhancement, getting feature sequence $F \in \mathbb{R}^{T \times (HW+K) \times h}$. Firstly, we utilize each query token to calculate different patch-level spatial clues via the attention mechanism for each frame:
\begin{equation}
    x_i = \text{softmax}(q_i\cdot F^T)F,
\end{equation}
where frame-wise representation $x_i \in \mathbb{R}^{T \times h}$. We conduct a similar process with our object trajectory encoder to add a $\left[\mathtt{CLS}\right]$ token to the start 
 of $x_i$. Finally, we utilize the Spatial-Temporal Action Encoder to model the temporal clue across frames. The projected $\left[\mathtt{CLS}\right]$ output $x \in \mathbb{R}^{M \times d}$ is expected to encode spatial-temporal features needed for modeling the action feature. To better guide the training procedure, we propose a spatial-temporal action set prediction task, which will be illustrated in the following section.
\subsubsection{Modality-agnostic Encoder}
Given encoded frame-level video representation $v=\{v_1, v_2, \cdots, v_{T}\}$, we first do mean-pooling over the entire representation, getting video representation $v_{[\mathtt{CLS}]}$. All video features $v$ are then projected into the same projection dimension $d$. The final concatenated video representation is $v=\{v_1, v_2, \cdots, v_{T}, v_{[\mathtt{CLS}]}\}$, along with the above mentioned text feature $t$, temporal-aware object feature $o$ and spatial-temporal action feature $x$. The final concatenated feature is $[v, o, x, t] \in \mathbb{R}^{(T + 1 + N + M + L) \times d}$, which will be input to the modality-agnostic encoder to do a further cross-modal interaction. We take the output of $v_{[\mathtt{CLS}]}$ token $v^{o}$ as the overall feature representation of the video, following \cite{radford2021learning} to take the output of $t_{\left[\mathtt{SEP}\right]}$ token $t^{o}$ as the overall text representation.

\begin{table*}
\centering
\begin{tabular}{llcccccccc} 
\toprule[1.5pt]& \multirow{2}{*}{PT data} & \multicolumn{4}{c}{MSR-VTT} & \multicolumn{4}{c}{MSVD} \\
 & & C & M & R & B-4 & C & M & R & B-4 \\ 
\hline
%OA-BTG & COCO & 46.9 & 28.2 & - & 41.4 & 90.6 & 36.2 & - & 56.9 \\
%MGSA & - & 47.5 & 27.6 & - & 42.4 & - & - & - & - \\
%POS+VCT & - & 49.1 & 29.7 & 62.8 & 42.3 & - & - & - & - \\
ORG-TRL & - & 50.9 & 28.8 & 62.1 & 43.6 & 95.2 & 36.4 & 73.9 & 54.3 \\

SemSynAN & - & 51.9 & 30.4 & 64.7 & 46.4 & 111.5 & 41.9 & 79.5 & 64.4 \\
APML & - & 52.2 & 30.3 & 63.6 & 43.8 & 108.3 & 39.2 & 76.2 & 58.0 \\
VNS-GRU & - & 53.0 & 29.9 & 63.4 & 45.3 & 121.5 & 42.1 & 79.7 & \textbf{66.5} \\
SAM-SS & - & 53.2 & 29.3 & 63.6 & 45.8 & 109.7 & 39.0 & 77.0 & 62.4 \\
SwinBERT & - & 53.8 & 29.9 & 62.1 & 41.9 & 120.6 & 41.3 & 77.5 & 58.2 \\
Clip4Caption& 136M & 57.7 & 30.7 & 63.7 & 46.1 & - & - & - & - \\
UniPerceiver & 45.4M & - & - & - & - & 131.0 & 42.3 & 79.0 & 61.5 \\
MV-GPT\dag & 53M & 60.0 & \textbf{38.7} & 64.0 & \textbf{48.9} & - & - & - & - \\
\hline
STOA-VLP& 2.5M & \textbf{60.2} & 31.0 & \textbf{68.4} & 45.8 & \textbf{131.8} & \textbf{43.4} & \textbf{83.9} & 64.4 \\
\toprule[1.5pt]
\end{tabular}
\caption{Video captioning results on the MSR-VTT and MSVD datasets. \dag: using subtitles as an additional input.}

\label{caption}
\end{table*}

\subsection{Pre-training Objectives}
\label{objective}
As shown in Figure \ref{model}, our STOA-VLP is pre-trained with four types of pre-training tasks, including two canonical types (i.e. overall video-text alignment and conditional language modeling). In overall video-text alignment, two instance-level pre-training tasks: video-text contrastive (VTC) loss and video-text matching (VTM) are utilized to build a global alignment between paired video and text. In Conditional language modeling, mask language modeling (MLM) and prefix language modeling (PLM) are incorporated to strengthen the modality-agnostic encoder's ability to generate fluent text conditioned on the given videos and texts. Besides those canonical objectives, we propose two new pre-training tasks---dynamic object-text alignment (OTA) and spatial-temporal action set prediction (ASP) to align temporal-aware objects with text and to learn to extract relevant spatial-temporal action features, respectively. We will further introduce these objectives next.
\subsubsection{Overall Video-Text Alignment}
VTC and VTM loss is used to learn an overall alignment between paired video and text. For modality-specific video and text encoders, we utilize 
symmetric contrastive mechanism (i.e. VTC) to encourage an early alignment:
\begin{equation}
    \mathcal{L}_{t 2 v}=-\sum_{i} \log \frac{\exp \left(\text{sim}\left(v^i, t^i\right)\right)}{\sum_{j} \exp \left(\text{sim}\left(v^j, t^i\right)\right)},
\end{equation}
\begin{equation}
    \mathcal{L}_{v 2 t}=-\sum_{i} \log \frac{\exp \left(\text{sim}\left(v^i, t^i\right)\right)}{\sum_{j} \exp \left(\text{sim}\left(v^i, t^j\right)\right)},
\end{equation}
\begin{equation}
    \mathcal{L}_{\text{VTC}} = (\mathcal{L}_{t 2 v} + \mathcal{L}_{v 2 t}) / 2,
\end{equation}
where $v^i$ is the video representation $v_{[\mathtt{CLS}]}$ for sample $i$, $t^i$ is the text representation $t_{[\mathtt{sep}]}$ for sample $i$. Note that the $\text{sim}(\cdot)$ operation is based on the cosine distance. Then, following \cite{li2021align}, hard negative video-text combinations that are similar with each true video-text pairs is sampled using similarity matrix $\text{sim}(v, t)$ compute for VTC loss. The VTM is applied to distinguish the original video-text pairs from constructed ones. Therefore a late alignment is built after the modality-agnostic encoder.

\subsubsection{Conditional Language Modeling}
We utilize conditional language modeling tasks to further enhance the modality-agnostic encoder's conditional text generation ability. %following \cite{devlin2018bert}, 
We randomly mask 15\% of input text tokens and predict them over the whole vocabulary.

We follow \cite{wang2021simvlm} to apply the PLM task, which masks a random length suffix of the given text, and tries to predict the mask place with the prefix text left.

\subsubsection{Dynamic Object-Text Alignment}
To further enhance the alignment between the extracted object trajectory information and the paired text, we propose a dynamic object-text alignment. The idea is to align temporal-aware object trajectories with relevant objects from the text. We use SpaCy\footnote{https://spacy.io/} to apply POS-tagging on all texts. Text encoder's output of text tokens tagged as nouns are denoted as $t_{\text{noun}} \in \mathbb{R}^{L_{\mathtt{noun}} \times d}$. We find a optimal bipartite matching between extracted noun tokens $t_{\text{noun}}$ and temporal-aware object trajectory $o_t$ based on the Hungarian algorithm \cite{kuhn1955hungarian} to find an optimal alignment between object trajectories and relevant noun tokens. We assume the found optimal matching is the contextualized ground truth alignment, and optimize the alignment with the following equation:
\begin{equation}
    [o_{\text{ind}}, t_{\text{ind}}] = \text{matching}(\text{Softmax}(\text{sim}(o_t, t_{\text{noun}}))),
\label{matching}
\end{equation}
\begin{equation}
    \mathcal{L}_{\text{OTA}} = \text{MSE}(o_{\text{ind}}, t_{\text{ind}}),
\end{equation}
where $|o_{\text{ind}}| =  |t_{\text{ind}}| = \text{min}(K, L_{\mathtt{noun}})$. $o_{\text{ind}}, t _{\text{ind}}$ represents matching tokens from $o_t$ and $t_{\text{noun}}$, respectively. The overall text representation $t_{[\mathtt{CLS}]}$ is included in $t_{\text{noun}}$. Being matched with the $t_{[\mathtt{CLS}]}$ represents the corresponding object trajectory fails to be matched with any specific noun token. Note that the text encoder is frozen during this process. By optimizing this objective, the object trajectory encoder should produce a spatial-temporal aware feature that connects closer to the paired text.
\subsubsection{Spatial-Temporal Action Set Prediction}
Instead of carefully labeling action sequences for the video which might not be relevant to the text due to the transfer of domains, a more flexible way to guide the spatial-temporal action modeling procedure is to predict action out of verb tokens from the paired text. To achieve this goal, we extract tokens tagged as verbs similar to OTA. The action sequence from text usually is not in chronological order, since humans tend to summarize and reconstruct information from the video. Thus, we follow \cite{carion2020end} to define a action set $s_{\text{act}}=\text{Set}(t_{\text{act}}) \cup \{\emptyset\}$. An optimal matching can be calculated with equation (\ref{actmatching}) between action feature $x$ and action set $s_{\text{act}}$. The matching result is viewed as the prediction ground truth. The prediction loss is calculated as follows:
\begin{equation}
    [x_{\text{ind}}, s_{\text{ind}}] = \text{matching}(\text{Softmax}(\text{sim}(x_t, s_{\text{act}}))),
\label{actmatching}
\end{equation}
\begin{equation}
    \mathcal{L}_{\text{ASP}} = \text{MSE}(x_{\text{ind}}, s_{\text{ind}}).
\end{equation}
To optimize the $\mathcal{L}_{\text{ASP}}$, encoded action tokens and matched verbs tend to be closer. Note that when the action token is matched with the $\emptyset{}$ (here we use the overall text representation, i.e. $\emptyset{} = t_{[\mathtt{CLS}]}$), the predicted action is none.
\subsubsection{Overall Objective}
The final loss function of our proposed STOA-VLP is:
\begin{equation}
    \mathcal{L} =\mathcal{L}_{\text{VTC}} + \mathcal{L}_{\text{VTM}} + \mathcal{L}_{\text{MLM}} + \mathcal{L}_{\text{PLM}} + \mathcal{L}_{\text{OTA}} + \mathcal{L}_{\text{ASP}}.
\end{equation}
By incorporating two spatial-temporal objectives, which are complementary to the overall alignment. Our STOA-VLP learns a better coarse-to-fine cross-modal connection, thus benefiting the downstream tasks that require a more detailed understanding of the video and text. 

\begin{table*}
\centering
\begin{tabular}{llccccccccc} 
\toprule[1.5pt]
 & \multirow{2}{*}{PT data} & \multicolumn{3}{c}{MSRVTT} & \multicolumn{3}{c}{LSMDC} & \multicolumn{3}{c}{DiDeMo} \\
 &  & R@1 & R@5 & R@10 & R@1 & R@5 & R@10 & R@1 & R@5 & R@10 \\ 
\hline
Frozen\ddag & 5.5M & 32.5 & 61.5 & 71.2 & 15.0 & 30.8 & 40.3 & 31.0 & 59.8 & 72.4 \\
ALPRO\ddag & 5.5M & 33.9 & 60.7 & 73.2 & - & - & - & 35.9 & 67.5 & 78.8 \\
VIOLET & 11.5M & 34.5 & 63.0 & 73.4 & 16.1 & 36.6 & 41.2 & 32.6 & 62.8 & 74.7 \\
HDVILA & 103M & 35.6 & 65.3 & 78.0 & 17.4 & 34.1 & 44.1 & 28.8 & 57.4 & 69.1 \\
OA-Trans\ddag  & 5.5M & 35.8 & 63.4 & 76.5 & 18.2 & 34.3 & 43.7 & 34.8 & 64.4 & 75.1 \\
BridgeFormer\ddag & 5.5M & 37.6 & 64.8 & 75.1 & 17.9 & 35.4 & 44.5 & 37.0 & 62.2 & 73.9 \\
\hline
\textit{CLIP initialization} \\
OA-Trans$\ddag \diamond$ & 5.5M & 40.9 & 70.4 & 80.3 & - & - & - & - & - & - \\
CLIP4CLIP & 380K & 44.5 & 71.4 & 81.6 & 21.6 & 41.8 & 49.8 & 43.4 & 70.2 & 80.6 \\
CAMOE* & - & 44.6 & 72.6 & 81.8 & 22.5 & 42.6 & 50.9 & 43.8 & 71.4 & - \\
BridgeFormer$\ddag \diamond$ & 5.5M & 44.9 & 71.9 & 80.3 & 21.8 & 41.1 & 50.6 & - & - & - \\
CLIP2Video & - & 45.6 & 72.6 & 81.7 & - & - & - & - & - & - \\
CLIP2TV* & - & 48.3 & 74.6 & 82.8 & - & - & - & - & - & - \\
MDMMT-2\dag & 140M & 48.5 & 75.4 & \textbf{83.9} & \textbf{26.9} & \textbf{46.7} & 55.9 & - & - & - \\ 
\hline
STOA-VLP& 2.5M & \textbf{50.1} & \textbf{75.5} & 83.8 & 24.8 & 46.2 & \textbf{56.0} & \textbf{51.1} & \textbf{76.4} & \textbf{84.0} \\
\toprule[1.5pt]
\end{tabular}
\caption{Text-video retrieval comparison results on MSR-VTT, LSMDC, and DiDeMo datasets, *: selecting results without post-processing, e.g. dual softmax \cite{cheng2021improving}. \dag: using additional training corpus. \ddag: pre-training with the same Web-Vid2M dataset, plus an image-text dataset CC3M\cite{sharma2018conceptual}. We neglect CC3M  because it's hard to model object trajectories and actions with only one image. $\diamond$: additional results via initializing the model from CLIP.} 
\label{retrieval1}
\end{table*}

\section{Experiment}
\subsection{Pre-training Datasets}
Instead of pre-training on the commonly used dataset HowTo100M \cite{miech2019howto100m} with 136M noisy pre-training data which only contains instructional videos, we utilize the WebVid-2M \cite{bain2021frozen} to prevent an enormous computation cost. Webvid-2M contains 2.5M web-collected video-text pairs and is an order of magnitude less than the HowTo100M.

\subsection{Implementation Details} 
Both the single-modal video and language encoder are 12-layer transformers \cite{vaswani2017attention}. We initialize the parameters from  (ViT-B/16) \cite{radford2021learning}. Similar to \cite{li2022align}, The 6-layer modality-agnostic fusion encoder is initialized with the first 6 layers of the CLIP Text Encoder. The 2-layer object trajectory encoder and spatial-temporal action encoder are initialized from the top 2 layers of the CLIP ViT. We uniformly sample 12 frames from each video. We resize and center crop them into 224x224 to split into patches with size 16x16, getting H=W=14. The maximum length of the text is set to 32. We select K=10 objects per frame. We set the number of object trajectory tokens to 20, and the number of action trajectory tokens is set to 4. Inspired by \citet{cordonnier2019relationship}, the 10th and 11th layer's patch output of the video encoder is used to generate object trajectories and action features, respectively. All pre-trainings are conducted on 128 Nvidia Tesla V100 GPUs with a batch size of 1024.

\subsection{Results on Downstream tasks}
We evaluate STOA-VLP on three commonly used video-text understanding and generation tasks, i.e. video captioning, text-video retrieval, and video question answering. 

\subsubsection{Video Captioning}

We compared our STOA-VLP with recent works including task-specific methods: ORG-TRL \cite{zhang2020object}, SemSynAn \cite{perez2021improving}, APML \cite{lin2021augmented}, VNS-GRU \cite{chen2020delving}, SAM-SS \cite{chen2020semantics}, SwinBERT \cite{lin2022swinbert}; and pre-training methods: Clip4Caption \cite{tang2021clip4caption}, UniPerceiver \cite{zhu2022uni} and MV-GPT \cite{seo2022end}.

We conduct experiments on the commonly used MSR-VTT and MSVD datasets. Table \ref{caption} shows the results of the following four metrics: BLEU-4 (B-4) \cite{papineni2002bleu}, CIDEr (C) \cite{vedantam2015cider}, METEOR (M) \cite{banerjee2005meteor} and ROUGE-L (R-L) \cite{lin2004rouge}. Our proposed STOA-VLP surpasses other methods on most of the metrics, even when pre-trained with less video-text data.

\begin{table}[htbp]
\centering
\begin{tabular}{llccc} 
\toprule[1.5pt]
    & \multirow{2}{*}{PT data} & MSR-VTT & MSVD & MC\\ 
    & & Acc & Acc & Acc\\
\hline
ActBERT &136M & - & - & 48.6 \\
NoiseE &136M & 35.1 & 35.1 & - \\
HCRN& - & 35.6 & 36.1& - \\
ClipBERT&5.6M & 37.4 & - & 88.2\\
VLM&136M&-&-& 91.6 \\
VideoClip &136M&-&-& 92.1 \\
DeCEMBERT&136M & 37.4 & - \\
CoMVT& 35M& 39.5 & 42.6& - \\
HDVILA&103M & 40.0 & - & 97.1\\
JustAsk& 69M& 41.5 & 46.3& - \\
MV-GPT&53M & 41.7 & - \\
ALPRO\ddag&5.5M & 42.1 & 45.9& - \\
Merlot&180M & 43.1 & - & 90.9\\
VIOLET&185.5M & \textbf{43.9} & 47.9 & 91.9\\

\hline
STOA-VLP& 2.5M& 43.2 & \textbf{50.8}& \textbf{98.5} \\
\bottomrule[1.5pt]
\end{tabular}
\caption{Video question answering results on MSR-VTT-QA, MSVD-QA, and MSR-VTT MC datasets. \ddag: pre-training with the same Web-Vid2M dataset, plus an image-text dataset CC3M.}
\label{qaaa}
\end{table}

Although MV-GPT utilizes subtitles as an additional input, our model still significantly outperforms it on Rouge-L(+4.4) and CIDEr(+0.2) without any additional text input during fine-tuning. The steady improvement demonstrates that STOA-VLP can better utilize both the coarse video representation and the fine-grained trajectory information to generate consistent and fluent captions.

\subsubsection{Text-Video Retrieval}

 We list Recall at Rank K ($R@K$) results of Frozen \cite{bain2021frozen}, ALPRO \cite{li2022align}, VIOLET \cite{fu2021violet}, HDVILA \cite{xue2022advancing}, OA-Trans \cite{wang2022object}, BridgeFormer \cite{ge2022bridging} and those initialize weights from CLIP \cite{radford2021learning}: CLIP4CLIP \cite{luo2021clip4clip}, CAMOE \cite{cheng2021improving}, CLIP2Video\cite{fang2021clip2video}, CLIP2TV \cite{gao2021clip2tv}, MDMMT-2 \cite{kunitsyn2022mdmmt}, and STOA-VLP on MSR-VTT-1k-A \cite{xu2016msr}, LSMDC \cite{anne2017localizing} and DiDeMo \cite{rohrbach2015long}.

 Table \ref{retrieval1} shows the results. Though not explicitly enhancing the overall alignment, our proposed method surpasses most previous methods, except for MDMMT-2 which uses additional training corpora of 10 downstream text-video datasets. We observe that initializing weights from CLIP \cite{radford2021learning} boosts the performance of retrieval. Our STOA-VLP surpasses other pre-training methods initialized from CLIP \cite{radford2021learning}, especially the OA-Trans and the BridgeFormer, which combine the Web-Vid2M we used and the CC3M during pre-training. This further demonstrates that dynamically modeling object trajectories and action features can also build a fine-grained connection that benefits the overall video-text alignment.

\begin{table*}
\centering
\begin{tabular}{lccccccccccccc} 
\toprule[1.5pt]
\multirow{2}{*}{} & \multirow{2}{*}{Obj.} & \multirow{2}{*}{Act.} & \multirow{2}{*}{\begin{tabular}[c]{@{}c@{}}OTA \\\end{tabular}\begin{tabular}[c]{@{}c@{}}OTA \\\end{tabular}} & \multirow{2}{*}{ASP} & \multicolumn{4}{c}{MSR-VTT Captioning} & \multicolumn{4}{c}{MSVD Captioning} & MSVD QA \\
 &  &  &  &  & C & M & R & B-4 & C & M & R & B-4 & Acc \\ 
\hline
Base &  &  &  &  & 58.3 & 30.6 & 68.2 & 44.8 & 127.3 & 42.1 & 82.7 & 62.3 & 49.6 \\ 
\hline
 & \checkmark &  &  &  & 58.7 & 30.8 & 68.2 & 45.2 & 128.7 & 42.3 & 83.0 & 62.7 & 50.1 \\
 & \checkmark &  & \checkmark &  & 59.6 & 30.7 & 68.2 & 45.1 & 131.7 & 42.9 & 83.4 & 63.0 & 50.7 \\
 & \checkmark & \checkmark & \checkmark &  & 58.9 & 30.9 & 68.3 & 44.9 & 128.0 & 43.3 & 83.5 & 63.6 & 50.3 \\ 
\hline
STOA-VLP & \checkmark & \checkmark & \checkmark & \checkmark & \textbf{60.2} & \textbf{31.0} & \textbf{68.4} & \textbf{45.8} & \textbf{131.8} & \textbf{43.4} & \textbf{83.9} & \textbf{64.4} & \textbf{50.8} \\
\toprule[1.5pt]
\end{tabular}
\caption{The ablation of our proposed modules, We add our proposed modules and pre-training tasks in sequence to evaluate their effectiveness on video captioning and video question answering. Note that Obj. means the object trajectory tokens, Act. means the spatial-temporal action tokens.}
\label{detablation}
\end{table*}

\subsubsection{Video Question Answering}

We evaluate our model on MSR-VTT-QA \cite{xu2017video}, MSVD-QA \cite{xu2017video} and MSR-VTT-MC \cite{yu2018joint} datasets with the prediction accuracy metric. We compare our results with task-specific method HCRN \cite{le2020hierarchical} and pre-training methods: ActBERT \cite{zhu2020actbert}, NoisE \cite{amrani2021noise}, ClipBERT \cite{lei2021less}, VLM \cite{xu2021vlm}, VideoClip \cite{xu2021videoclip}, DeCeMBERT \cite{tang2021decembert}, CoMVT \cite{seo2021look}, HDVILA \cite{xue2022advancing}, JustAsk \cite{yang2021just}, MV-GPT \cite{seo2022end}, ALPRO \cite{li2022align}, Merlot \cite{zellers2021merlot} and VIOLET \cite{fu2021violet}. 

Table \ref{qaaa} shows the results. With only 2.5M pre-training data, our model surpasses all other methods on MSVD-QA, except for VIOLET which pre-trains on a 185.5M dataset and incorporates DALL-E \cite{ramesh2021zero} pre-trained on 250 million text-image pairs to generate discrete visual features for learning. We observe a performance gain of 2.9\% on MSVD-QA, 1.4\% on MSR-VTT-MC compared to the previous best models. We conjecture that by explicitly modeling object trajectories and actions, a better alignment is built between the question and the visual feature, and fine-grained information from video is observed and utilized to better answer text questions.

\begin{figure}[h]
\centering
\subfigure[\# action tokens]{
\includegraphics[width=1.6in]{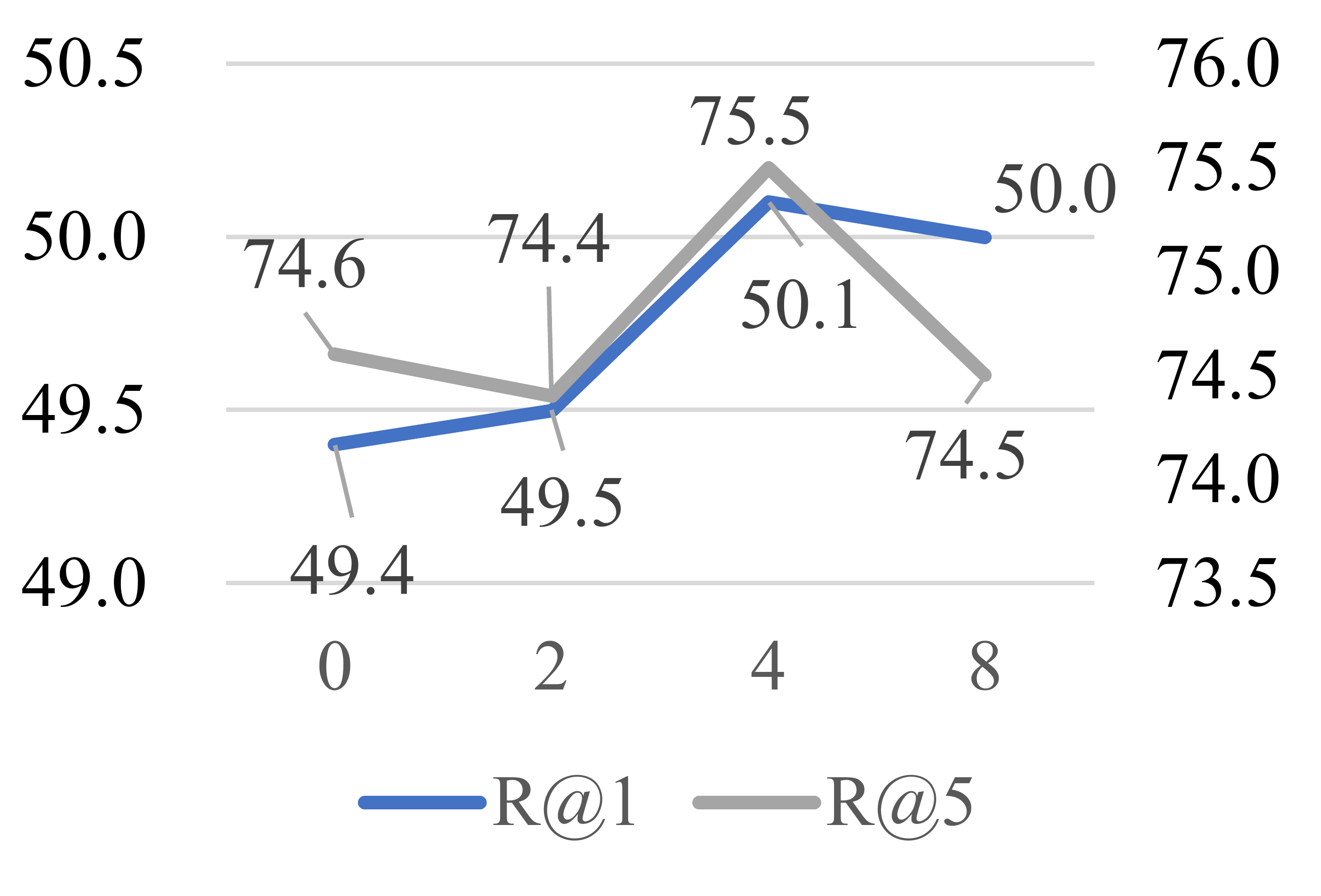}}
\subfigure[\# object trajectory tokens]{
\includegraphics[width=1.6in]{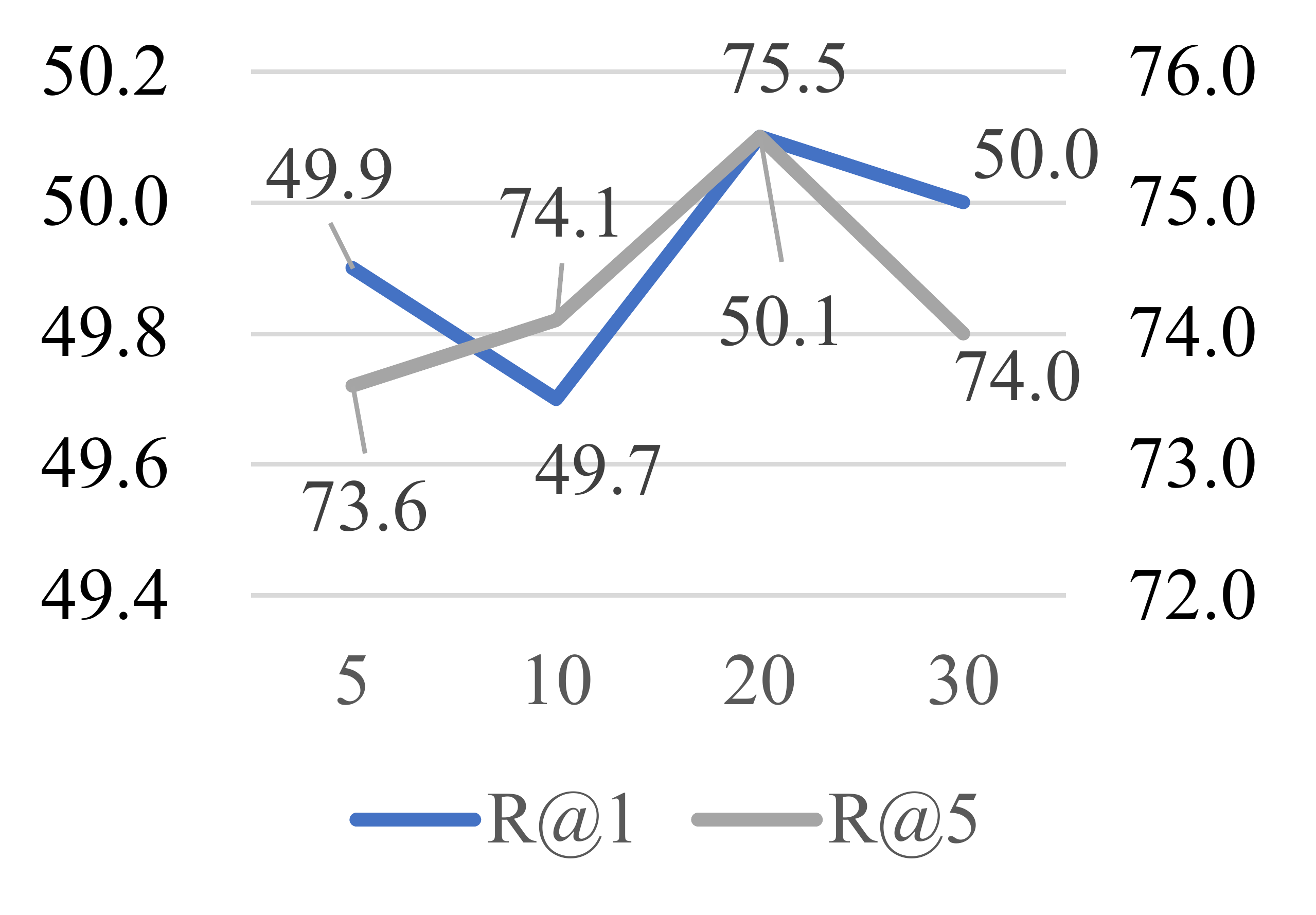}}

\caption{(a): the number of object trajectory tokens and the corresponding retrieval results. (b): the number of action tokens and the corresponding retrieval results. Note that the action token number is set to 4 for (a), and the object trajectory token number is set to 20 for (b).}

\label{ablation}
\end{figure}

\subsection{Ablation Studies}
We conduct ablation studies on the MSR-VTT text-video retrieval task and the MSR-VTT, MSVD video captioning task to evaluate the effects of different components.
\subsubsection{Effect of Numbers of Spatial-temporal Tokens}
We study the effect of the number of temporal-aware object trajectory tokens and spatial-temporal action tokens on the MSR-VTT retrieval task. 
As shown in Figure \ref{ablation} (a), we find that the incorporation of the action token does help to align between video and text. Generating 4 action tokens yields the best retrieval performance. Continue to enlarge the action token number, we observe a performance drop. We conjecture this is because some of the action tokens are redundant and failed to be matched with the verb tokens in the text. We found a similar pattern for varying the number of object trajectory tokens in Figure \ref{ablation} (b).

\subsubsection{Effect of Proposed Components}
We further analyze the effect of our incorporated features and proposed tasks to model them on two different downstream tasks including video captioning and video question answering, which naturally require more fine-grained information and a spatial-temporal understanding of the video scene. The results are shown in Table \ref{detablation}. Note that the base model removes all spatial-temporal modules and auxiliary modeling tasks. Compared to the base model, the incorporation of temporal-aware object trajectories alone can bring improvements to all tasks. Our OTA task further builds a fine-grained alignment between object trajectories and noun tokens that benefits the downstream tasks. We also find that adding a spatial-temporal action modeling module without the ASP task makes some of the captioning results worse. Video captioning and video question answering tasks require a fine-grained semantic understanding of the visual part. We conjecture that without the guidance of our ASP task, the meaning of our action tokens is ambiguous, which might cause a performance drop. Together with our proposed spatial-temporal modules and tasks, our best result is achieved.

\section{Conclusion}
In this paper, we explicitly model fine-grained information in a spatial-temporal way to build a better cross-modal alignment. Our proposed STOA-VLP introduces two novel modules to model object trajectories and action features across space and time dimensions. Two auxiliary tasks are designed to build a coarse-to-fine cross-modal alignment. We observe advanced improvements in downstream tasks with moderate-size pre-training data.
In addition, the proposed framework can also be applied to more tasks, we regard it as a future working direction.

\newpage
\section{Acknowledgments}
Xiaocheng Feng is the corresponding author of this
work. We thank the anonymous reviewers for their insightful comments. This work was supported by the National Key R\&D Program of China via
grant 2020AAA0106502, National Natural Science Foundation of China (NSFC) via grant 62276078
and the Major Key Project of PCL, PCL2021A06.

\bibliography{stoa_vlp_arxiv}

\part*{\quad \quad \ \ Technical Appendix}
\section{Introduction}
In this technical appendix, we include:
\begin{itemize}
    \item More implementation details for pre-training
    \item More downstream task details
    \item Additional ablation results
    \item Case study
\end{itemize}
\section{More Implementation Details}
In our STOA-VLP, the visual representation dimension $h$ is set to 768, and the cross-modal representation dimension $d$ is set to 512. The $M$ action query tokens $q = \{q_1, q_2, ..., q_M\}$ are randomly initialized, where $q_i \in \mathbb{R}^{1 \times h}$. During the pre-training stage, we adopt Adam \cite{kingma2014adam} as the optimizer with an initial learning rate of 1e-5, betas of (0.9, 0.98). The cosine schedule \cite{loshchilov2016sgdr} scheme is adopted to decay the learning rate. We pre-train our STOA-VLP for 15K steps in total. 16 Nvidia Tesla V100 GPUs are used during the fine-tuning stage.
\section{Downstream Task Details}
\subsection{Datasets}
\begin{itemize}
    \item \textbf{MSR-VTT} \cite{xu2016msr} contains 10,000 videos, with 20 human-annotated descriptions for each video. A publicly adopted split \cite{chen2019motion} is used for video captioning, where 6,513/497/2,990 videos if used for training/validation/test. The training-9k split \cite{gabeur2020multi} is adopted for text-video retrieval. In video question answering, split \cite{xu2017video} is adopted.
    \item \textbf{MSVD} \cite{chen2011collecting}  contains 1,970 videos with 80K descriptions. For video captioning, the dataset is split into 1,200, 100, and 670 videos for training, validation, and testing. For video question answering, the split used in \cite{xu2017video} is adopted, where 1,200, 250, and 520 videos are used for training, validation, and testing, respectively. 
    \item \textbf{LSMDC} \cite{rohrbach2015long} contains 118,081 video clips from 202 movies. For text-video retrieval, according to \cite{fang2021clip2video}, the validation set and the test set contain 7,480 and 1,000 videos, the remaining videos are for training.
    \item \textbf{DiDeMo} \cite{anne2017localizing} contains 10,000 Flickr videos with 40K descriptions. For this dataset, we follow \cite{bain2021frozen, luo2021clip4clip} to concatenate all descriptions for one video as a single query to perform paragraph-video retrieval. 
\end{itemize}
\subsection{Video Captioning}
Video captioning requires the model to generate a relevant description given the video, which describes what presents and happens in the corresponding video. For this task, object trajectories and action tokens are generated with encoded video patch features. During the training stage, a lower-triangle attention mask is applied to the text embedding, ensuring that only visual features and previous words can be seen when predicting the current word. During the inference, A fixed-length mask sequence is utilized to generate the caption auto-regressively with mask tokens, in which the model generates one word at a time. We fine-tune our STOA-VLP for 5 and 10 epochs on MSR-VTT and MSVD, respectively. 

Among all the compared methods, 
task-specific methods ORG-TRL \cite{zhang2020object}, SAM-SS \cite{chen2020semantics}, and SemSynAn \cite{perez2021improving} are designed to better incorporate fine-grained semantic features. Multiple decoders are applied for the APML \cite{lin2021augmented}. \citet{chen2020delving} further investigate the decoder and propose three techniques to improve the performance of VNS-GRU. In SwinBERT \cite{lin2022swinbert}, video swin transformer \cite{liu2022video} is utilized to extract sparse video tokens from densely sampled video frames. The end-to-end pre-training and fine-tuning for captioning is enabled in Clip4Caption \cite{tang2021clip4caption} and MV-GPT \cite{seo2022end}. In UniPerceiver \cite{zhu2022uni}, a unified architecture is designed to process visual and text information concurrently.

\subsection{Text-video Retrieval}
Text-video retrieval requires the model to find the paired video from multiple candidate videos when provided with the query text. We compute matching probabilities via the modality-agnostic fusion encoder. Firstly, similar to the VTC task, We utilize the outputs from the video encoder and text encoder to calculate the similarities $p_{v^i}$ between the query text and all video candidates via the following equation: 
\begin{equation}
    p_{v^i}= \log \frac{\exp \left(\text{sim}\left(v^i, t\right)\right)}{\sum_{j} \exp \left(\text{sim}\left(v^j, t\right)\right)},
\end{equation}
where $t$ is the query text's embedding, $v^i$ is the candidate video's embedding. For a higher computation efficiency, we select the 32 most similar examples via $p_{v^i}$ to obtain the matching probability. Subsequently, with the object trajectories and action features generated from the encoded video patch features, the final text-video matching probability is obtained via the modality agnostic fusion encoder. During the fine-tuning stage, we apply the VTC and VTM tasks. Note that we do not apply post-processing methods like dual softmax \cite{cheng2021improving} and QB-Norm \cite{bogolin2022cross}. We fine-tune our STOA-VLP for 5 epochs on all retrieval datasets. 

Among all the compared methods, 
Frozen \cite{bain2021frozen}, OA-Trans \cite{wang2022object}, CLIP4CLIP \cite{luo2021clip4clip}, CAMOE \cite{cheng2021improving}, BridgeFormer \cite{ge2022bridging}, \cite{fang2021clip2video}, and MDMMT-2 \cite{kunitsyn2022mdmmt} adopt a dual-encoder only structure for an efficient retrieval. A video-text matching module is built on top of the dual encoder structure in the CLIP2TV \cite{gao2021clip2tv}. The fusion encoder with multiple pre-training tasks is introduced to further align video and language modality in VIOLET \cite{fu2021violet}, HDVILA \cite{xue2022advancing}, and ALPRO \cite{li2022align}.

\subsection{Video Question Answering}
Video question answering (Video QA) aims at understanding the text question, as well as preserving relevant fine-grained information from videos to answer the question. We formulate the task as an answer classification problem. The video feature, object trajectories, action tokens, and the corresponding question with a single additional answer mask token are fed into the modality agnostic fusion encoder and predict the answer token within the whole vocabulary. The Multiple-Choice (MC) VideoQA requires the model to select from several candidate options rather than answering questions, therefore we do it the same way with the video-to-text retrieval task. We fine-tune our STOA-VLP for 5 epochs on all three datasets. 

Among all compared methods, a conditional relation network is introduced in the task-specific method HCRN \cite{le2020hierarchical} to construct more sophisticated structures for representation and reasoning. The rest are all pre-training methods. Modality agnostic encoder and modality agnostic pre-training paradigm are introduced in VLM \cite{xu2021vlm}. It's worth noting that JustAsk \cite{yang2021just} is pre-trained on a large-scale question answering dataset HowToVQA69M \cite{yang2021just}.  The connection between visual content and the text is important for video question answering, as well as the temporal alignment. Similar to MV-GPT \cite{seo2022end}, next utterance prediction is used in CoMVT \cite{seo2021look} to better incorporate visual context. Temporal alignment is further explored in NoisE \cite{amrani2021noise}, VideoClip \cite{xu2021videoclip}, DeCeMBERT \cite{tang2021decembert}, VIOLET \cite{fu2021violet} and Merlot \cite{zellers2021merlot}. 

\section{Additional Ablation study}
\begin{figure}[h]
\centering
\includegraphics[width=2.3in]{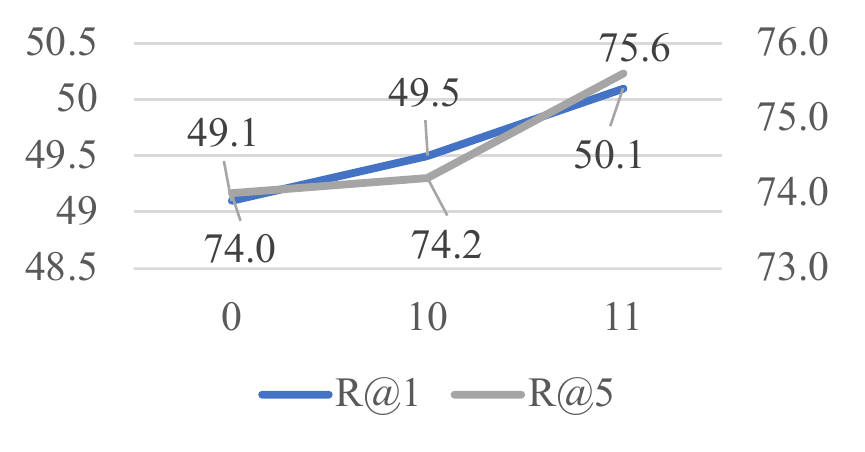}
\caption{The layer of the object patch feature $O^\tau$ is extracted from}
\label{layer}
\end{figure}

Different transformer layers can focus on the different regions of the visual features. Lower layers focus more on the local patterns, while higher layers focus more on the larger patterns \cite{cordonnier2019relationship}. Therefore, on the MSR-VTT text-video retrieval task, We perform an additional ablation study on the layer of the object patch feature $O^\tau$ is extracted from, which results in a different semantic representation for extracted object features. Note that we extract features from the output of the layer.
From figure \ref{layer}, it can be seen that extracting object patch features and generating object trajectories from the deeper layers benefits the retrieval task more than the lower layers (e.g. from the output of layer 0, raw video patches are utilized). We conjecture that object features extracted from deeper layers preserve more high-level semantic information which is more helpful for modeling object trajectories and action features.

\section{Case study}
\begin{figure*}[t] 
\centering
\includegraphics[width=2.0\columnwidth]{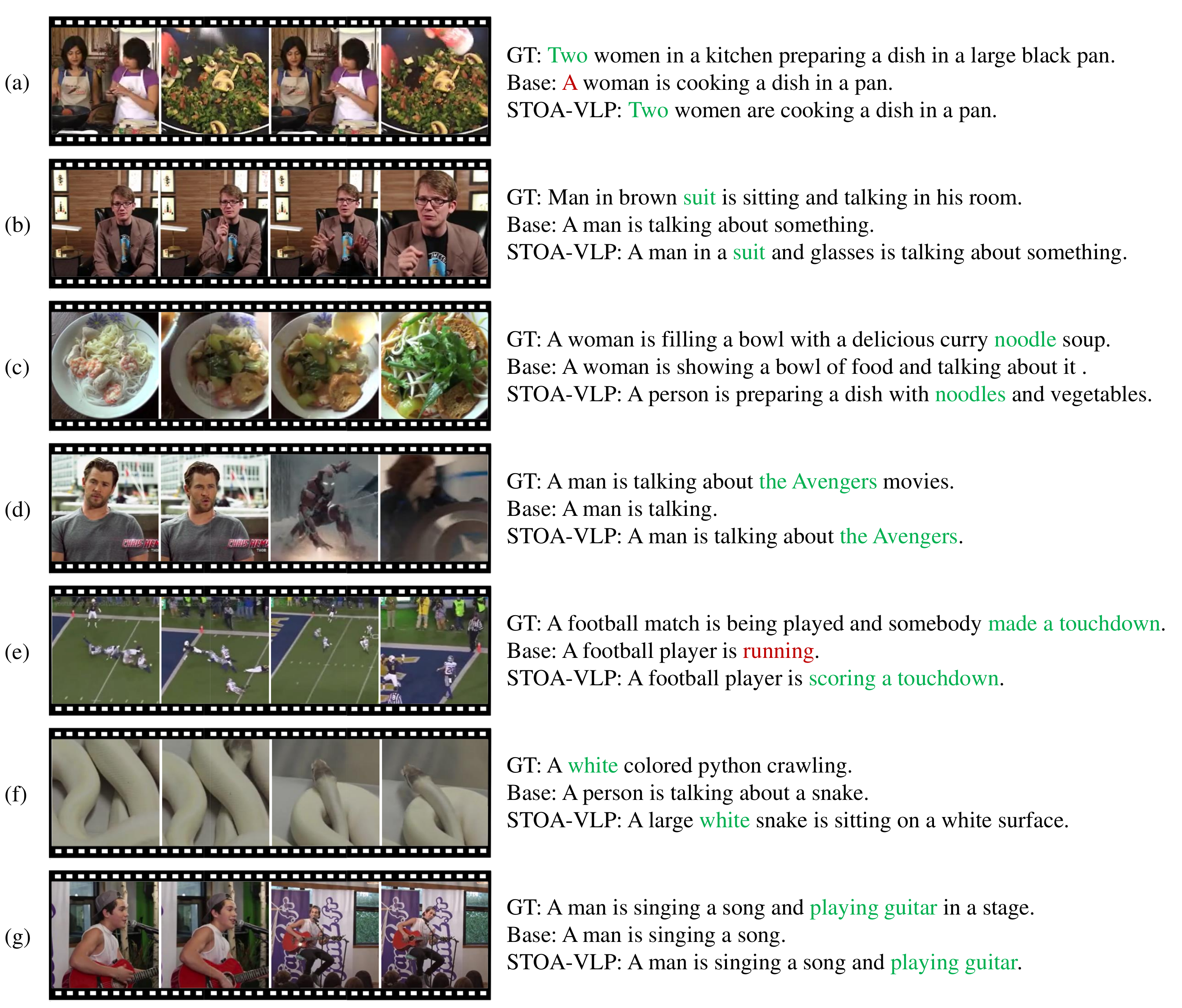}
\caption{Video captioning examples generated by our STOA-VLP and the base model. Compared to the base model, texts generated by STOA-VLP that are more consistent with the ground truth are marked in green. Texts generated by the base model that are inconsistent with the ground truth are marked in red.}
\label{case}
\end{figure*}

In this section, we show some video captioning examples generated by our STOA-VLP, compared to our base model, from which our object trajectory encoder and spatial-temporal action encoder are removed, as well as the OTA and ASP tasks. The examples are shown in Figure \ref{case}.

From Figure \ref{case}(a) we can observe that our STOA-VLP correctly recognizes the number of women presenting in the video. More fine-grained details are described by STOA-VLP in Figure \ref{case} (b) and (c). We conjecture that this is because our STOA-VLP can better recognize the objects, as well as preserve more fine-grained features with the help of incorporated object trajectories and action features.
It can be seen from Figure \ref{case}(d) that our STOA-VLP can better connect different scenes from the video and generate consistent captions. Our model even recognizes the man is talking about 'the Avengers' based on frames that show relevant movie characters. 
Our STOA-VLP also can better understand the main contents of the video. In Figure \ref{case} (e) and (f), our model concentrates more on the actions that humans tend to focus on, which might be because our action encoder is trained with the ASP task to generate action features that are aligned with the actions in the paired text. What's more, in Figure \ref{case}(h), we notice that STOA-VLP can generate multiple actions triggered by different object interactions, further demonstrating that modeling multiple actions is useful.

%With the proposed framework, it can also be applied to more tasks like action localization and recognition, which can improve the action token generation procedure in turn, if applied to the pre-train stage, we leave it as a future working direction.

\newpage

% Use \bibliography{yourbibfile} instead or the References section will not appear in your paper
\bibliography{stoa_vlp_arxiv}

\end{document}